\documentclass{article}

\usepackage{arxiv} 

\usepackage[utf8]{inputenc} % allow utf-8 input
\usepackage[T1]{fontenc}    % use 8-bit T1 fonts
\usepackage{hyperref}       % hyperlinks
\usepackage{url}            % simple URL typesetting
\usepackage{booktabs}       % professional-quality tables
\usepackage{amsmath,amsfonts,amssymb}       % blackboard math symbols
\usepackage{nicefrac}       % compact symbols for 1/2, etc.
\usepackage{microtype}      % microtypography
\usepackage{graphicx}

\usepackage[backend=biber, style=numeric, sorting=none, citestyle=numeric]{biblatex}
\usepackage{doi}
\usepackage{multirow}

\title{Compressed Meta-Optical Encoder for Image Classification}

\author{Anna Wirth-Singh$^{1,*}$\\
\And
Jinlin Xiang$^{2,*}$\\
\And
Minho Choi$^{2,*}$\\
\And
Johannes E. Fr\"{o}ch$^{1,2}$\\
\And
Luocheng Huang$^{2}$\\
\And
Shane Colburn$^{2}$\\
\And
Eli Shlizerman$^{3,2}$\\
\And
Arka Majumdar$^{1,2}$\\
\And
\\
$^{1}${Department of Physics, University of Washington, Seattle, WA 98195, USA}\\
$^{2}${Department of Electrical and Computer Engineering, University of Washington, Seattle, WA 98195, USA}\\
$^{3}${Department of Applied Mathematics, University of Washington, Seattle, WA 98195, USA}\\
\\
$^{*}${These authors contributed equally to this work. }
}

\date{April 2024}

\begin{document}
\maketitle

\begin{abstract}  
Optical and hybrid convolutional neural networks (CNNs) recently have become of increasing interest to achieve low-latency, low-power image classification and computer vision tasks. However, implementing optical nonlinearity is challenging, and omitting the nonlinear layers in a standard CNN comes at a significant reduction in accuracy. In this work, we use knowledge distillation to compress modified AlexNet to a single linear convolutional layer and an electronic backend (two fully connected layers). We obtain comparable performance to a purely electronic CNN with five convolutional layers and three fully connected layers. We implement the convolution optically via engineering the point spread function of an inverse-designed meta-optic. Using this hybrid approach, we estimate a reduction in multiply-accumulate operations from 17M in a conventional electronic modified AlexNet to only 86K in the hybrid compressed network enabled by the optical frontend. This constitutes over two orders of magnitude reduction in latency and power consumption. Furthermore, we experimentally demonstrate that the classification accuracy of the system exceeds 93\% on the MNIST dataset.
\end{abstract}


%%%%%%%%%%%%%

\section{Introduction}\label{sec1}
Convolutional neural networks (CNNs) represent a significant milestone in image classification, recognition, and tracking~\cite{krizhevsky2012imagenet}. CNNs, for example, AlexNet, are composed of several convolutional layers that adaptively learn spatial representations from input images. While powerful, the convolution operation is computationally expensive, leading to high latency and power consumption. In fact, it has been estimated that about 80\% of the total runtime of CNNs is used in performing convolution operations \cite{Li2016}. Reducing this latency and as a result power consumption has become an active area of research, with multiple works proposing free-space optical systems as a solution \cite{McMahon2023, Chang2018, Colburn2018, Yang2023}. Beyond latency reduction and power consumption, optical information processing features qualities including high bandwidth, spatial parallelism, and low-loss transmission which have led to a surge of interest in the field \cite{McMahon2023}.  

For decades, it has been known that a $4f$ lens system can be used to perform convolutions optically by placing an appropriate filter at the Fourier plane of the lens \cite{Cutrona1960,Mait2018,Burgos21,Chang2018}. This was demonstrated in 2018 \cite{Chang2018} using a diffractive optical element as the filtering element and traditional refractive lenses composing the $4f$ system. Spatial light modulators \cite{Yang2023} and digital micromirror devices \cite{Hu2022} can also be used as the filtering element. However, one drawback of the Fourier-based $4f$ approach is that it requires three elements (two lenses and a spatial filter), resulting in a bulky optical system with greater propensity for misalignments than single-element optical systems. Such misalignments from each optical convolutional layer cannot be ignored even when weights are trained with noisy inputs (more details in the Supplementary material). In addition, the filtering optics must be contained within a compact area at the focal plane in the $4f$ system, which limits parallel processing ability unless creative measures are taken such as utilizing naturally present diffraction orders \cite{Hu2022} or lenslet arrays \cite{Colburn2018}.

Advantageously, the convolution operation can also be performed using free-space optics and requires only a single element. The resultant image produced by any optics is the input convolved with the point spread function (PSF) of the optics \cite{Lohmann1978,Mait1986,Mait1987}. Therefore, by engineering optics to produce a particular PSF, convolution can be performed optically simply via passing light through the optics. Further, by passing the input through several of these optics in parallel, multiple convolution operations can be performed simultaneously at the speed of light \cite{Colburn2018,Wei2023}. This approach leverages the inherent parallelism of light enabling the passive processing of a vast amount of data without increasing computation time \cite{McMahon2023,Hu2022,Colburn2018}. This unique optical capability circumvents scalability issues when handling high-resolution images in traditional electronic-based CNN systems. 

However, a challenge to all optically-implemented CNN approaches is that nonlinear layers are interspersed with the linear layers. For example, AlexNet consists of five convolutional layers followed by three fully connected layers~\cite{krizhevskyimagenet}. Specifically, each convolutional layer in the architecture utilizes the rectified linear unit (ReLu) as its non-linear activation function, followed by the local response normalization and Max Pooling layer, ensuring an effective mechanism for spatial hierarchy extraction. Therefore, nonlinearity is consistently applied across all layers, serving as a foundational element of the network's design to enhance its learning capability. Without the particular nonlinear layers that are effective in CNNs (e.g., ReLu), the classification accuracy of the CNN drops by about 20\% \cite{xiang2022knowledge}. The nonlinear layers cannot be implemented using simple lens-like optics; to implement them optically, some physical nonlinearity must be introduced, for instance by using an atomic vapor cell \cite{Yang2023,Ryou2021} or image intensifier \cite{Wang2023}. Hybrid approaches involving repeated transduction of the signal to perform linear operations in optics and nonlinear operations in electronics provide little benefit due to large latency and power consumption in signal transduction \cite{Colburn2018,Wang2023,Burgos21}. Implementing only one of many required convolution operations does not provide much benefit in terms of speed and latency. Alternatively, there have been recent breakthroughs in using end-to-end designs for physical and hybrid networks designs which perform image classification or other tasks without explicitly using convolution  \cite{Zheng2022,McMahon2023,Zhou2021,Huang2023}. Such an approach can effectively implement multiple linear layers in one optical frontend. While novel, these end-to-end neural networks are computationally expensive to train and are applicable only to the physical system for which they were specifically designed.

\begin{figure}[h!]
\centering\includegraphics[width=15cm]{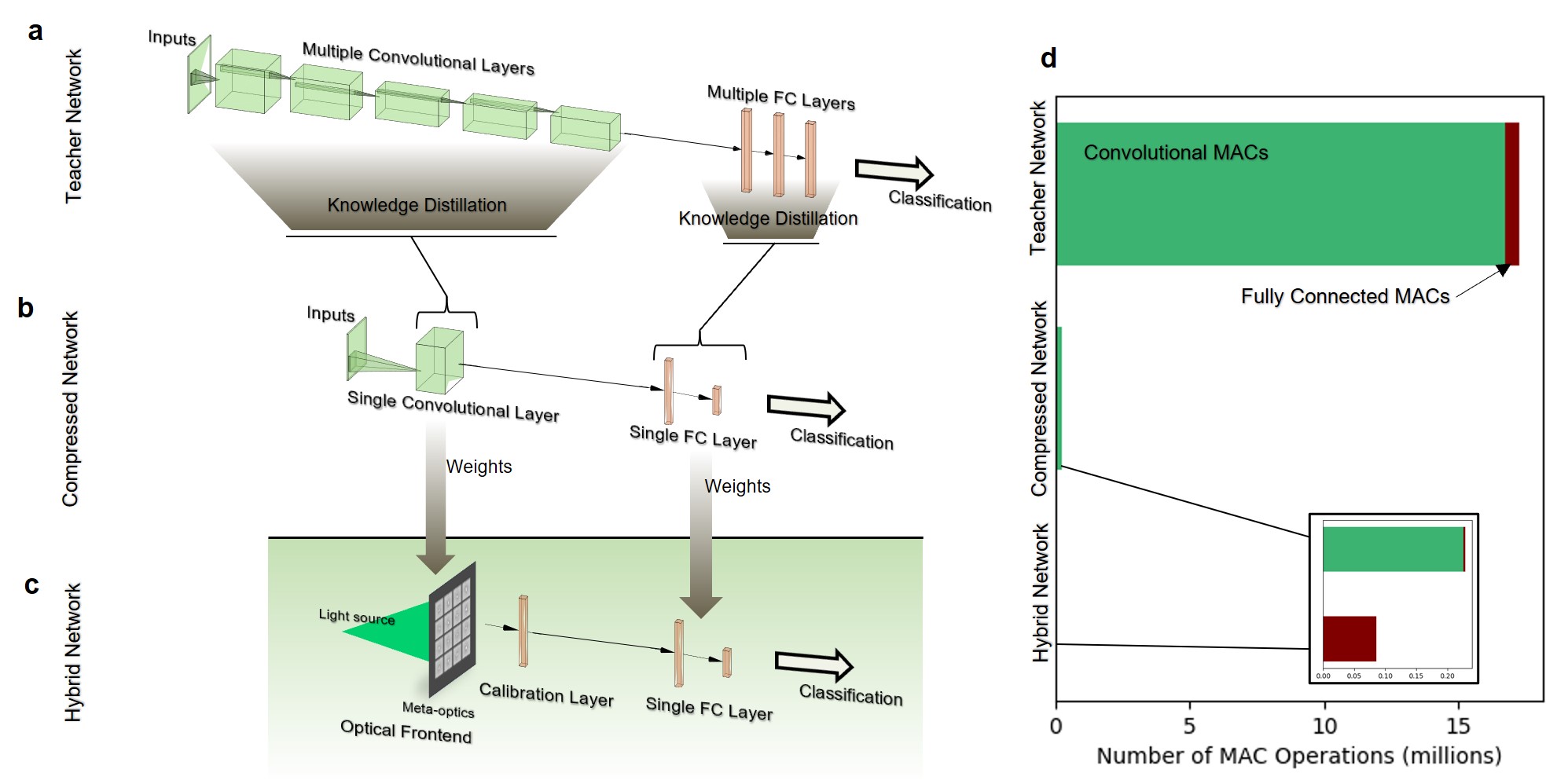}
\caption{Schematic of convolutional neural networks for image classification tasks. (a) All-electronic multi-layered CNN. (b) All-electronic compressed CNN. (c) Hybrid CNN which combines an optical meta-optic front end and electronic backend. (d) The number of multiply-accumulate (MAC) operations of each network configuration, with convolutional MACs in green and fully-connected (FC) MACs in brown.}
\label{Fig:CNN_Schematic}
\end{figure}

In this work, we experimentally demonstrate a hybrid optical-electronic CNN consisting of a single optical convolution layer with an electronic single fully connected layer to achieve similar accuracy as AlexNet on hand-written digit classification tasks. To overcome the limitation from the absence of optical nonlinearity, we apply knowledge distillation to remove the nonlinear layers, and compress multiple layers into a single linear layer~\cite{xiang2022knowledge}. Knowledge distillation (more details in Methods) circumvents the need for nonlinearity without a significant reduction in the performance by transferring knowledge from a larger, pre-trained network (the `teacher' network) to a more compact network (the `student' network). Here, we use a modified AlexNet, denoted AlexNet-Mod, as the teacher network and a single convolutional layer coupled with a single fully connected layer as the student network. We use this architecture to demonstrate a hybrid meta-optical platform, wherein an optical frontend based on a single meta-optic performs the linear convolution operation, followed by an electronic backend which contains a linear calibration layer and a fully connected layer. In such a way, the most computationally expensive operation is performed optically to leverage the benefits of optical computing, namely high spatial bandwidth and low power consumption. The use of a single meta-optic layer drastically simplifies the experimental setup and provides a compact geometry.

The optical frontend of this network is realized using inverse-designed meta-optics. The meta-optics are arrays of sub-wavelength scatterers which act as phase masks, imparting spatially-coded phase shifts to incident light. Here, we design meta-optics to realize a phase mask which performs the desired convolutional steps of the CNN by engineering the PSF. We fabricate and experimentally validate the performance of the designed optics using incoherent green light illumination from a light emitting diode, centered at 525 nm. Further, we experimentally demonstrate the classification accuracy of the entire hybrid CNN on the MNIST dataset. The hybrid CNN is described in Figure \ref{Fig:CNN_Schematic}, where we compare the architectures of multi-layer electronic CNNs, compressed electronic CNNs (a linear single layer CNN), and our hybrid system. The number of multiply-accumulate (MAC) operations in the entire network is reduced by over two orders of magnitude through compressing multiple convolutional layers into a single layer and implementing them optically. The classification accuracy of the compressed hybrid CNN is reduced by only 5\% from AlexNet-Mod (98\% accuracy) to achieve 93\% classification accuracy on the MNIST dataset.

\section{Results}\label{sec2}

In each convolutional layer of a CNN, an optimized kernel is convolved with the input to generate a feature map which is then passed to the next layer. Using the knowledge distillation approach, we optimize eight convolutional kernels (each $6 \times 6$ pixels in size) for the MNIST dataset of handwritten digits. The selected number and size of the kernels are based on previous experimental results (details shown in the Supplementary Information)~\cite{ozturk2018convolution}. As described in the next section, we design the optics to implement these optimized convolutional kernels and combine them with an electronic backend for image classification.

\subsection{Compressing Multiple Convolutional Layers using Knowledge Distillation}\label{subsec11}

To select the ideal network architecture for the linear optical-electronic hybrid system, underfitting and overfitting issues must be avoided. Smaller networks face underfitting concerns, particularly in optical settings that limit the system to just 8 kernels and remove nonlinear functions~\cite{cai2021network}. This is in stark contrast to AlexNet, which uses over 300 kernels~\cite{krizhevsky2012imagenet}. However, complex models are prone to overfitting, and practical issues such as fabrication noise and misalignments introduce further challenges for optically implementing a large number of kernels. 

Therefore, to obtain a balanced network that could be implemented optically, we use knowledge distillation to compress a AlexNet-Mod as a base model. AlexNet-Mod consists of 5 convolutional layers and 3 fully connected layers (8 total layers) which we compress to the desired structure of one convolutional layer and one fully connected layer (2 layers). The knowledge distillation approach assumes that the teacher network is already trained and performs the desired task with high accuracy; in this case, we use AlexNet-Mod as the teacher network, which achieves 98.9\%$\pm$ 0.33\% on training and 98.4\%$\pm$0.32\% on testing classification accuracy MNIST datasets over repeated trials. Additionally, AlexNet-Mod employs nonlinear activation functions (ReLU) to optimize performance; these are circumvented by knowledge distillation for a result that is compatible with our optical setting. In the compressed network, we limit the number of kernels in the compressed convolutional layer to 8, and each kernel is $6 \times 6$ pixels in size. After training, the compressed electronic network achieves an approximate classification accuracy of 96\% on both training and testing datasets. 

\begin{figure}[h!]
\centering\includegraphics[width=15cm]{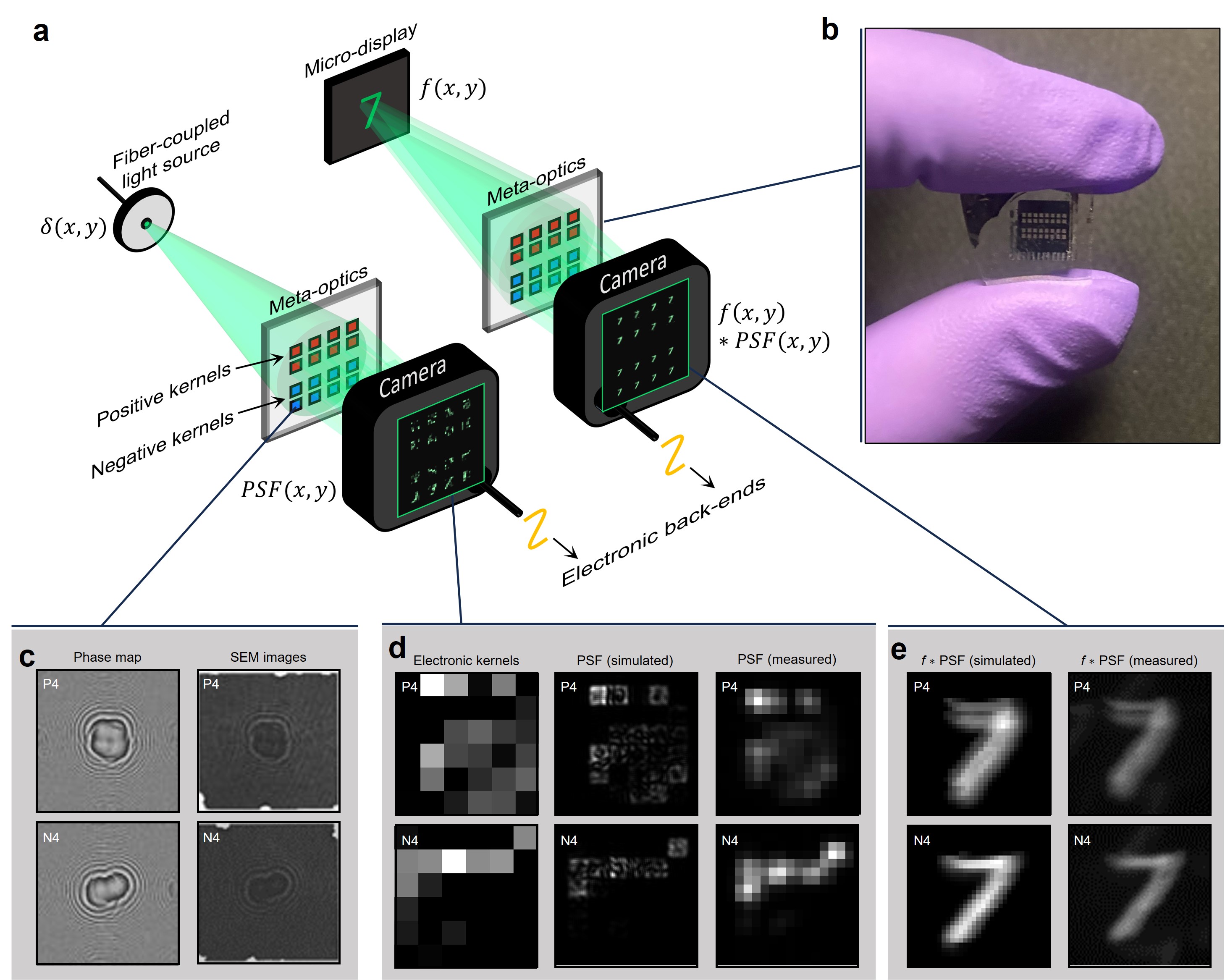}
\caption{Schematic of the optical system. (a) PSF measurement setup using a monochromatic point light source (left) and optical convolution measurements using a micro-LED display (right). (b) A photograph of the fabricated meta-optics. The meta-optic contains 16 different sub-optics, spatially distributed in a single layer, operating in parallel for classification tasks. (c) Phase maps and SEM images of exemplary sub-optics corresponding to the positive and negative parts of a particular convolutional kernel. (d) The positive and negative parts of an example convolutional kernel (left) and the corresponding PSF simulation (middle) and right (experiment). (e) The simulated electronic output (left) and optical experiment (right) convolved output for the example kernel, for the case of an input ``7" from MNIST. }
\label{Fig:Experiment_Schematic}
\end{figure}

\subsection{Optical Convolution using Meta-Optics}\label{subsec12}

The optical component of this network is implemented using inverse-designed meta-optics. We used the Gerchberg-Saxton (GS) phase retrieval method \cite{Gerchberg1972} to design phase masks that correspond to the optimized convolutional kernels. Specifically, each sub-optic is designed to have a PSF which resembles the convolutional kernel. Since electronic convolutional kernels include both positive and negative values, we separated each kernel into positive and negative parts and designed meta-optics for each. The positive and negative images were computationally subtracted afterward to produce the net convolution. Two-dimensional phase maps of an example set of sub-optics are shown in Fig. \ref{Fig:Experiment_Schematic}(b). The phase maps were implemented by silicon nitride pillars which are 750 nm tall but varied widths corresponding to their relative phase delay. Scanning electron microscope (SEM) images of the fabricated optics are also shown for comparison, highlighting the fabrication quality. All sixteen optics (corresponding to 8 convolutional kernels) were fabricated on a single substrate (more details in Materials and Methods), allowing convolution from 16 different kernels in a single image capture. A photograph of the fabricated meta-optic is shown in Fig. \ref{Fig:Experiment_Schematic}(b). Each kernel meta-optic is of size 470 $\mu$m $\times$ 470 $\mu$m and center-to-center distance of 705 $\mu$m. The optics are arranged in two rows of eight optics for a total footprint of 5.6 mm $\times$ 1.4 mm. 

To verify the performance of the optics, we measured the PSF of each sub-optic using a single-mode fiber as a point light source, as shown in the Fig.  \ref{Fig:Experiment_Schematic}(a). The PSFs from all 16 sub-optics are simultaneously captured by the two-dimensional CMOS camera (more details in Materials and Methods). Accordingly, the convolved images from all 16 sub-optics are also captured at the same time when we replaced the single mode fiber with the display, as described in Fig.~\ref{Fig:Experiment_Schematic}(a). Exemplary electronic convolutional kernels, which represent the ground truth PSFs for the optically-implemented kernels, are shown in Fig. \ref{Fig:Experiment_Schematic}(d). We also present the simulated PSFs from the meta-optics using angular spectrum propagation \cite{Goodman1996,Matsushima2009}. The experimentally measured PSF shown in Fig. \ref{Fig:Experiment_Schematic}(d) match well to the simulated PSFs, confirming fabrication accuracy. However, due to the constraints on physically realizable PSFs, there are notable differences between the ground truth PSFs and experimentally measured PSFs. To correct for these differences, as well as slight noise and misalignments in the optical system, we introduce a calibration layer to the computational backend, further discussed in Section \ref{subsec42}.

\subsection{Hybrid Network Classification Results}\label{subsec13}

To address optical noise and misalignment affecting image classification performance, an additional calibration layer is introduced to adjust optical representations for compatibility with the computational backend. Specifically, this calibration layer was fine-tuned with only 10\% of the training dataset and ensures that the computational backend does not need to be retrained. Therefore, the electronic backend of the hybrid network consists of the calibration layer followed by the original compressed electronic backend. 

We compare three CNN architectures (AlexNet-Mod, compressed electronic network, and hybrid optical-electronic network) in Table \ref{Tab1}. The AlexNet-Mod achieves classification accuracy exceeding 98\% on both training and testing datasets. The number of MAC operations of this network is 17 million with 8 bit precision (details are shown in the Supplementary Information, Section S1). The compressed electronic CNN achieves greater than 96\% accuracy; this 2\% decline reflects the inherent challenges of compressing multiple layers into a single layer. Primarily due to the compression of the convolution layers, the number of MAC operations is reduced to 228,672. The hybrid network, which integrates the optical convolution layer with the calibration layer and single fully connected layer electronic backend, experimentally achieves classification accuracy of 93.9\% (± 0.25\%) and 93.4\% (± 0.22\%) on the training and testing datasets, respectively, and requires only 85,824 MAC operations, which is 0.5\% and 37\% of that required for AlexNet-Mod and the compressed electronic networks, respectively. 

\begin{table}[h]
\caption{Classification Results}\label{tab2}
\label{Tab1}
\begin{tabular*}{\textwidth}{@{\extracolsep\fill}cccc} 
\toprule
Network Architecture & Train (\%) & Test (\%) & MAC Operations \\
\midrule
AlexNet-Mod & 98.9 ± 0.33 & 98.4 ± 0.32 & 17,268,224 \\
Compressed electronic CNN (without KD) & 84.2 ± 0.47 & 82.1 ± 0.69 & 228,672 \\ 
Compressed electronic CNN (KD) & 97.2 ± 0.35 & 96.2 ± 0.29 & 228,672 \\
\midrule
Hybrid CNN (KD) & 93.9 ± 0.25 & 93.4 ± 0.22 & 85,824 \\ 
\bottomrule
\end{tabular*}
\end{table}

% \begin{tabular*}{\textwidth}{@{\extracolsep\fill}lcccccc}
% \toprule%
% & \multicolumn{2}{@{}c@{}}{Electronic CNN} & \multicolumn{1}{@{}c@{}}{Hybrid CNN} \\\cmidrule{2-4}\cmidrule{5-7}%
% Network Architecture & AlexNet-Mod & Compressed & Compressed  \\
% MAC Operations & 688,092,160 & 228,672 & 85,824 \\
% \midrule
% Train Accuracy  & $98.9\% \pm 0.33\%$ & $97.2\% \pm 0.35\%$ &  $93.9\% \pm 0.25\%$ \\
% Test Accuracy  & $98.4\% \pm 0.32\%$ & $96.2\% \pm 0.29\%$  & $93.4\% \pm 0.22\%$ \\
% \botrule
% \end{tabular*}
% \label{Tab1}
% \end{table}
  
\begin{figure}[h!]
\centering\includegraphics[width=12cm]{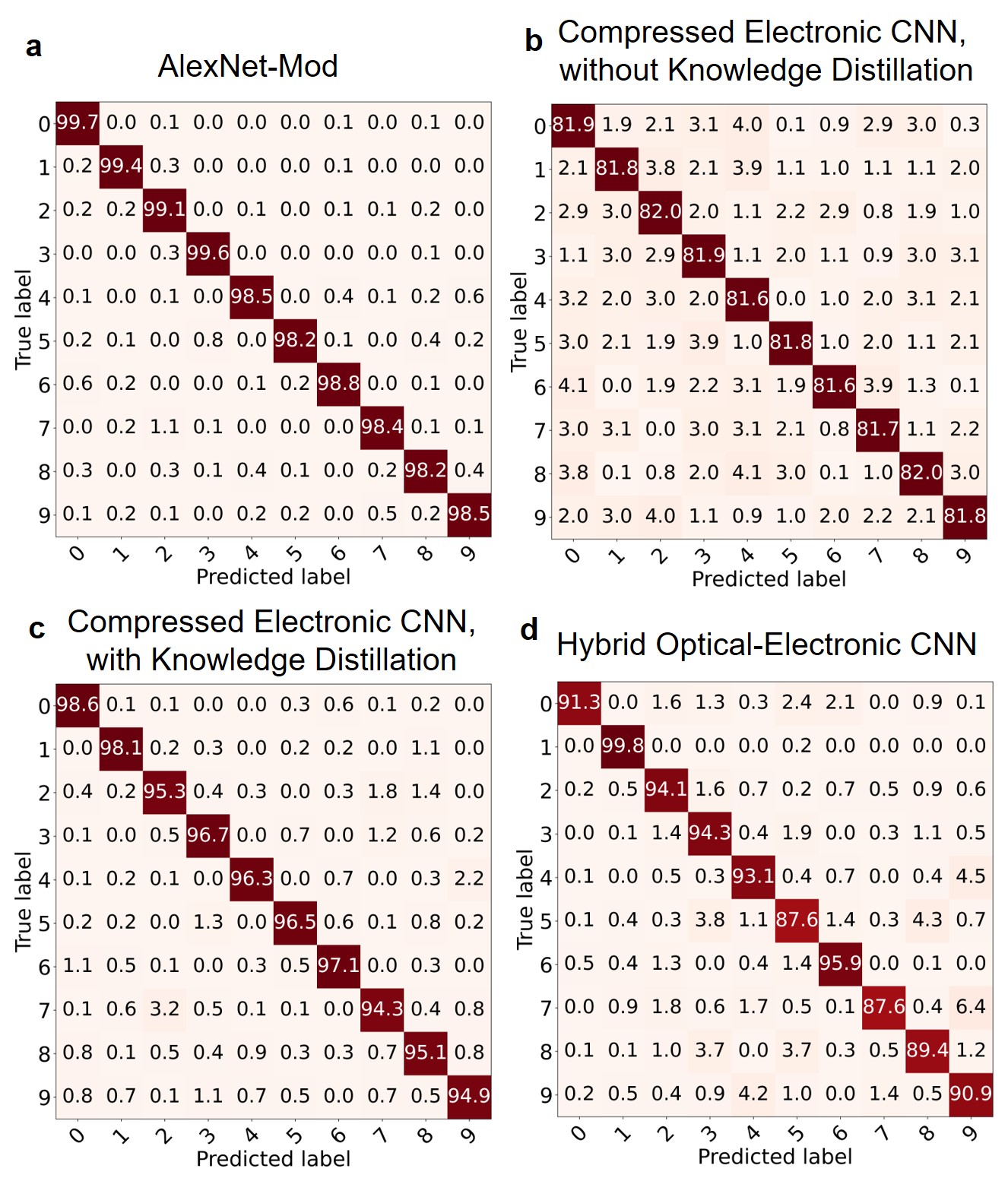}
\caption{Confusion matrices for different network architectures. (a) Classification results for AlexNet-Mod (multiple-layer electronic CNN). (b) Classification results for the all-electronic CNN compressed without using knowledge distillation. (c) Classification results for the all-electronic CNN compressed with knowledge distillation. (d) Classification results for the hybrid optical-electronic CNN.}
\label{Fig:ConfusionMatrices}
\end{figure}

Figure~\ref{Fig:ConfusionMatrices} illustrates the confusion matrices for three neural network configurations. Each matrix visually represents the model's tested performance across different classes (in this case, digits labeled 0 through 9), with the true labels on the rows and the predicted labels on the columns. The multi-layer electronic CNN, AlexNet-Mod, displays high values along the diagonal, exceeding 98.1\% accuracy on each class. The compressed electronic CNN, while having a slight decline in diagonal values, still demonstrates robust classification accuracy with a minimum of 94.3\%. The hybrid network exhibits a more diverse range of values along the diagonal, with some classes exhibiting lower predictive accuracy compared to the compressed electronic network. We attribute this slight decline to noise in the optical experiment, which may be due to optics fabrication, camera sensor noise, and optical noise due to vibrations that cannot be fully compensated by the calibration layer. Despite these factors, the hybrid network still performs reasonably well with the network correctly predicting each class with a minimum of 87.6\% accuracy. This indicates that the hybrid network maintains a reasonable level of accuracy against these noises and discrepancies. 

\section{Discussion}\label{sec4}

\subsection{Ablation Study and Principal Component Analysis}\label{subsec21}

To understand the contribution of each component of the hybrid optoelectronic CNN, we perform an ablation study and principal component analysis. In the ablation study, we evaluate the classification accuracy when using only the electronic backend structure for classification. We summarize the ablation study results in Table~\ref{tab: HNN_AS}. For a fair comparison, the backend layer here is the same structure as used in the hybrid network but re-optimized for the best performance in the absence of any convolutional frontend. Specifically, we compare the performance of a backend layer only (a single fully-connected layer), the calibration and backend layers together (two fully-connected layers), and the entire hybrid network. For a single fully-connected layer alone, an accuracy of 89\% was attained, which is less than that of the hybrid network; this highlights the utility of the optical frontend. Further, for two fully-connected layers (representing the calibration and backend layer) without any convolutional frontend, the accuracy is reduced to 84\%. This can be attributed to the fact that two layers are fully-connected and, in the absence of nonlinear activation functions, tend to converge towards a `saddle point' in the optimization landscape~\cite{jacot2021saddle}. 
\begin{table}[htpb]
\caption{Network Ablation Study}\label{tab: HNN_AS}
\begin{tabular*}{\textwidth}{@{\extracolsep\fill}ccc} 
\toprule
Configuration & Train accuracy & Test accuracy \\
\midrule
Backend Only & 89\% & 87\% \\
Calibration + Backend & 84\% & 80\% \\
Optics + Calibration + Backend & 94\% (+5\% / +10\%) & 93\% (+6\% / +13\%) \\
\bottomrule
\end{tabular*}
\end{table}

We hypothesize that the calibration layer only re-maps the optical representations and does not improve the performance of the electronic backend alone. To examine this, we further analyze the effect of the calibration layer and the overall performance of the hybrid network as compared to an all-electronic network using Principal Component Analysis (PCA). PCA is a statistical method widely used in various fields, especially in analyzing the quality of neural networks~\cite{ma2019dimension}, to project original, high-dimensional data into a new, simpler coordinate system for explicit interpretation. Specifically, PCA computes the eigenvalue decomposition of the covariance matrix or the singular value decomposition of input data to determine the principal direction (known as ``principal components"). Principal component 1 denotes the axis of maximum variance, encapsulating the most substantial relationships among variables, while principal component 2, orthogonal to principal component 1, captures the second most significant variance direction. Notably, the first two principal components typically contain the most crucial information. 
In this study, we compress the output dimensions from electronic and optical convolutional layers into two principal components, respectively, to compare the classification efficacy of each approach; this is observed by comparing the clusters observed in PCA visualizations~\cite{ivosev2008dimensionality}. 

\begin{figure}[h!]
\centering\includegraphics[width=15cm]{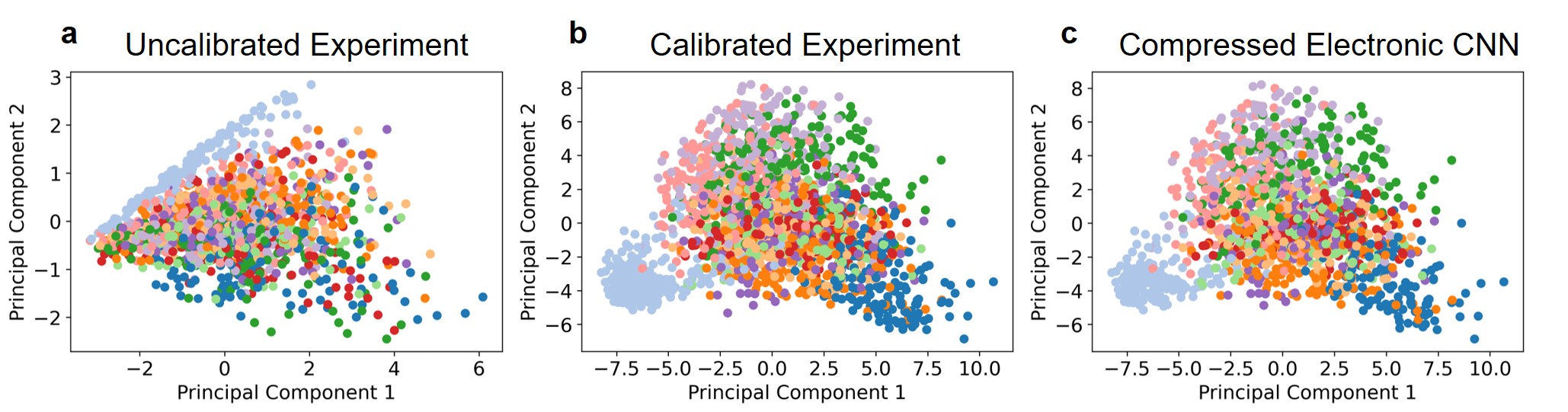}
\caption{PCA of the hybrid CNN. (a) PCA of the uncalibrated experimental hybrid CNN classification data. (b) PCA of the calibrated experimental data, which has been re-mapped and exhibits clustering behavior similar to that of the compressed electronic CNN data. (c) PCA of the compressed electronic CNN data.}
\label{Fig:PCA}
\end{figure}

In Fig.~\ref{Fig:PCA}, we use PCA to show that the raw experimental data do identify the fundamental components necessary for classification, but that the calibration function is necessary to shift optical representations to a form that is compatible with the pre-designed electronic backend. As shown in Fig. \ref{Fig:PCA}a and \ref{Fig:PCA}c, we observe that both the all-electronic CNN outputs and the uncalibrated hybrid network experimental results can effectively distinguish between different classes due to the clustering behavior of specific classes, e.g. light blue and navy blue. This clustering behavior indicates that despite any observable shifts in the PCA plot, the fundamental capacity of the hybrid network to classify data remains comparable to that of all-electronic networks. However, due to differences between the optical experiment output and the expected input to the electronic backend, directly using the original backend network results in a notable drop in accuracy, down to 16.3\%. The calibration layer is designed to re-calibrate the outputs from the optical convolution layer back to the original outputs, thereby enabling the use of the original backend without retraining. As shown in Fig. \ref{Fig:PCA}b and \ref{Fig:PCA}c, the calibrated experiment result exhibits very similar clustering behavior to the all-electronic network, further demonstrating that the hybrid network classification is comparable to that of the all-electronic network.

\subsection{PSF-Engineered Meta-Optics}\label{subsec22}

We emphasize two advantages of our PSF-engineering method to perform optical convolution. Firstly, we highlight the simplicity of the optical system, as this method requires only a display, a single layer of optic, and a camera, making it compact and simple to execute. Incoherent illumination is used, so this approach can be applied to real-world image classification scenarios. Secondly, we highlight the ease of integration with the optimized electronic system. One of the challenges faced by optical computing is that electronic computing is already extremely powerful, having had decades of research and development into algorithms and hardware \cite{McMahon2023}. In our approach, the optics are designed to implement the electronically-optimized kernels. These kernels may be modified to reflect improvements in electronic CNN models and architectures, and the optics can accordingly be adapted to implement convolutions with arbitrary kernel matrices. 

Furthermore, the Gerchberg-Saxon (GS) algorithm \cite{Gerchberg1972} used to design the optics is a well-established technique. The GS algorithm is an iterative phase retrieval algorithm to determine the phase (in the optic plane) that produces an intensity pattern in another desired plane (the focal plane). In other words, the meta-optics are phase-only holograms producing the desired PSFs as their images. Due to the fact that only amplitude, and not phase, contributes to the intensity pattern, the iteratively designed phase masks are not unique. More details on the implementation of the GS algorithm are available in the Supplement. In the Supplement, we also discuss an alternative design method based on automatic differentiation, which also produces viable optics but we found the GS-designed optics to produce slightly brighter, clearer images.  

There are, however, two major limitations of the PSF-engineering approach. One limitation is that there is no guarantee that the desired PSF is physically realizable. That is, a single phase mask that satisfies the desired amplitude constraints may not exist. However, by introducing an electronic calibration layer, the resultant PSF does not need to be perfect in order to effectively classify the data. Alternatively, to ensure physically realizable PSFs, one could adopt an end-to-end optimization scheme wherein the phase mask is simultaneously optimized with a backend. However, training this large phase mask (on the order of $10^{5} - 10^{6}$ unit cells per kernel optic) is prohibitively costly and the design space is potentially too large to attain convergence. In contrast, separately training the electronic convolutional kernels and the optics are both reasonable steps, which as demonstrated are effective when combined. 

A second limitation of the described approach is that we assume the PSF is spatially invariant. In reality, light from different spatial locations on the imaging object intersects the meta-optic at various angles of incidence, resulting in a different PSF for the off-axis rays. In contrast, we design the optics assuming a normal incident illumination. To mitigate this discrepancy, we ensure the incoming angles of incidence are relatively small by placing the display far away from the optics (90 mm) relative to the focal length (2.4 mm). Therefore, for a displayed image size of 8 mm $\times$ 8 mm, we ensure that the maximum deviation from normal incidence is $3.6^{\circ}$, and therefore the assumption of spatial invariance is reasonable for our system. However, for a large field of view imaging system, we may need to explicitly model the spatially varying PSF. We note that Wei et al. \cite{Wei2023} use reparameterization techniques to design spatially varying kernel optics and report higher classification accuracy using this method (73.8\%) versus designing optics with the assumption of spatial invariance (71.6 \%) on the CIFAR-10 dataset. In another variation of a PSF-engineering technique, Zheng et al. \cite{Zheng2024} engineer the PSF of polarization-sensitive meta-optics to provide an array of focal spots which produce images of intensities relative to the kernel weights.

\subsection{Outlook - Computational Effectiveness of the Hybrid Convolutional Neural Network}\label{subsec23}

The number of MAC operations required of a network serves as a metric of computational complexity which is independent of the employed hardware technology. In a modern digital system, one MAC operation consumes approximately $1 pJ$ \cite{Huang2023,Wang2023}, so the hybrid network is expected to reduce the required power to classify an input from $17 \mu J$ to $85 nJ$ based on the reduction in MAC operations. The latency of such a classification task is also expected to decrease proportionately. For input which is already in the optical domain, the power and latency required to capture an image and convert it to digital input is the same regardless of the network. Specifically, the hand-written digits of MNIST were captured using a standard camera, and the optical frontend of our network uses a standard camera sensor with meta-optics replacing the refractive camera lens. 

The benefit of optically implementing the convolutional step becomes more significant as the number of input pixels is increased. The electronic computational complexity of each convolutional layer is determined by the height and width of the input images ($H,W$), as well as the size of the kernels ($k$), and is $\mathcal{O}(HWk^2)$~\cite{xiang2022knowledge}. However, the computational complexity decreases to $\mathcal{O}(1)$ in the optical convolutional layer. For example, when the MNIST dataset's typical image size of 28$\times$28 pixels is increased to 100$\times$100 pixels, the electronic computational complexity of each convolutional layer is expected to increase \textbf{$12.76$} times (assuming the kernel size remains unchanged), but an optically implemented convolution would not incur any increase in computation time. Therefore, as the resolution of real-world images continues to increase, hybrid networks such as the one described offer a promising solution to scaling problems incurred by all-electronic networks. 

In summary, we demonstrate single-layer optical convolution with an electronic backend to achieve similar accuracy as AlexNet-Mod on MNIST hand-written digit classification, with 99.5\% reduction in computational complexity. To circumvent the nonlinearity of AlexNet-Mod, we use knowledge distillation to compress the CNN into linear layers which are then implemented in a hybrid format. As a further innovation, we implement the convolution optically via engineering the PSF of meta-optics, which results in a more compact and resilient optical frontend than the commonly used $4f$ lens system and does not require coherent illumination or polarization control. This hybrid approach integrates seamlessly with existing CNN architectures, utilizing simple optical design and requiring no re-training of the electronic backend to classify the data in experiment. This approach is also suitable for scaling to higher-resolution datasets; unlike in all-electronic networks where the convolution time scales with the number of input pixels, for optical convolution the processing time is independent of the resolution of the dataset. This work serves as a baseline for other optical and hybrid neural networks for higher bandwidth as well as lower power and latency in increasingly prevalent CNN applications.

\section{Materials and Methods}\label{sec5}

\subsection{Transfer knowledge to linear networks}\label{subsec41}

The knowledge distillation (KD) algorithm is designed to compress neural networks. KD accomplishes this by transferring knowledge from a larger, pre-trained network (referred to as the `teacher model') to a more compact network (referred to as the `student model'). In our implementation, we use AlexNet-Mod as the teacher network and a linear electronic network as the student network. The student network only comprises a single CNN coupled with a single fully connected layer. Straightforwardly training this linear network tends to easily converge to suboptimal saddle points; however, with the knowledge distillation approach, the student network converges faster and obtains better performance than it would achieve without the knowledge distillation training. 

The KD algorithm optimizes the linear student network by combining two types of losses: temperature loss and student loss. Similar to other conventional losses in image classification, KD uses the training labels as “hard labels” and computes the probabilities distribution vector 
$p^{hl}$, where each element $p_{i}^{hl}$
 in the vector corresponds to the probability of the current input belonging to the class $i$. The softmax function is used to compute probabilities 
\begin{equation}
    p_{i}^{hl}=exp(z_i)/\sum{exp(z_i)}
\end{equation}

where $z_i$ is the student logits after the last fully connected layer. 

Then, KD treats the prediction of the teacher model as `soft labels' that inform the student model since training with hard labels only is a sensitive process, especially for compact linear networks. KD includes a softening parameter, $T$, named as the distillation temperature for teacher probabilities. Therefore, for each input, at the same time of computing $p^{hl}$ with the student model, KD computes the soft probabilities vector, $p^{sl}$, with the teacher model according to 
\begin{equation}
    p_{i}^{sl}=exp(y_i/T)/\sum{exp(y_i/T)}
\end{equation}

Therefore, the total loss is then calculated as a weighted summation of the two losses
\begin{equation}
    \mathcal{L}(x,\Phi) = \alpha \mathcal{L}_C(y,p^{hl})+(1-\alpha)\mathcal{L}_k ((p^{sl,t};T = \tau),(p^{sl,s};T = \tau))
    \label{eq: loss}
\end{equation}

where $x$ corresponds to the input, $y$ is the training data, $\Phi$ are the student model weights, $\mathcal{L}_C$ is the cross-entropy loss function, $\mathcal{L}_k$ is the Kullback-Leibler (KL) Divergence Loss function~\cite{seo2020kl}, $p^{hl}$ corresponds to the student model hard predictions, $p^{sl,s}$ corresponds to the student predictions under given teacher model probabilities $p^{sl,t}$, and $\alpha$ is the weighting parameter.

\subsection{Calibrating Optical Experiment Results with Limited Data}\label{subsec42}

The calibration function is designed to remap the optical convolution outputs to align with those of the previously trained backend. This addition addresses a variety of differences which may occur between the optical and electronic counterparts, including scaling, translation, rotation, and optical noise. With the addition of the calibration layer, the weights of the fully connected layer are preserved, and the optical frontend can be integrated into the existing network framework without retraining the backend or fine-tuning the optical alignment. 

Specifically, the calibration layer is a fully connected layer. The loss function used in this process is defined:
\begin{equation}
    \mathcal{L} = min(f_{calibrate}(\textit{ON}),\textit{EN})
\end{equation}
where $ON$ is the network with the optical experiment results and $EN$ is the all-electronic network. This approach aims to refine the experiment output to align more closely with the pre-designed electronic network. To prevent overfitting, we strategically limit our training to only 10\% of the available data, ensuring that our model remains efficient~\cite{xiang2023tkil,zheng2020bi}. The calibration layer addresses diverse types of noise encountered in the optical system, thus ensuring a more robust hybrid network.

\subsection{Meta-optics Design}\label{subsec33}
The meta-optics are designed with our experiment setup in mind. Due to the sensitivity of our camera (GT-1930C) and available light sources, we design the optics specifically for 525 nm illumination. For all electromagnetic simulations, we use a simulation grid size of 586 nm to be both comparable to the wavelength of the light and evenly divisible by the size of the camera pixels (5.86 $\mu$m per pixel). Each sub-optic is square, 800x800 simulation pixels in size, which provide compact footprint and reasonable computation time. While it is not necessary to propagate the electric fields on a sub-wavelength grid, it is necessary to design the meta-optic scatterers with sub-wavelength periodicity. Therefore, we divide each meta-optic pixel into a 2x2 block of square meta-optic scatterers each with a period of 293 nm. With 750 nm SiN (n = 2.06) pillars, we select a set which provides 0 to 2$\pi$ phase shift at the desired wavelength. The scatterer unit cells were simulated using S4 RCWA \cite{Liu2012}. More details on the meta-optic design are available in the Supplement.

\subsection{Meta-optics Fabrication}\label{subsec34}
The convolutional meta-optics are fabricated on a silicon nitride layer on a quartz substrate. We first deposited silicon nitride on a double-side polished quartz wafer using plasma-enhanced chemical vapor deposition (Oxford; Plasma Lab 100). Then we patterned on a positive-tone resist (ZEP-520A) using e-beam lithography (JEOL; JBX6300FS). We used alumina as a hard mask for etching the silicon nitride layer, so we deposited the alumina using e-beam evaporator (CHA; SEC-600) and did liftoff with 1-methyl-2-pyrrolidinone. We etched the silicon nitride layer with a plasma etcher (Oxford; PlasmaLab 100, ICP-180) using fluorine-based gases. In order to minimize the stray light of the meta-optics, we blocked the light except for the 16 kernel sub-optics by putting apertures around each one using photolithography (Heidelberg; DWL66+) and metal deposition followed by a liftoff process.

\subsection{Optical Measurements}\label{subsec35}

Two-dimensional point spread functions of 16 different optical kernels are measured simultaneously with a simple measurement setup. Single-mode optical fiber-coupled light source acts as a point source, and the meta-optics are placed 92 mm apart from the source. As we put the meta-optics on a three-axis linear stage and a kinetic mount with two rotation adjusting knobs, both the position and angle of the meta-optics can be well-defined with respect to the designed setup. Then we put a high-resolution color camera (GT-1930C with 5.86 $\mu$m per pixel resolution) 2.4 mm from the meta-optics to collect the point-spread functions of the kernels. We simply replaced the light source from the single-mode optical fiber to a micro-display presenting MNIST dataset of handwritten digits to get the convolved images. The camera could capture all 16 convolved images from different kernels at the same time, and we used Python code to automatically collect convolved images from 10,000 number of MNIST dataset.

\section*{Supplementary Information}
The online version contains supplementary materials.

\section*{Code Availability}
The code and data generated and/or analyzed are available from the corresponding author upon reasonable request.

\section*{Acknowledgements}

The research is supported by National Science Foundation (NSF-ECCS-2127235 and EFRI-BRAID-2223495).  Part of this work was conducted at the Washington Nanofabrication Facility/ Molecular Analysis Facility, a National Nanotechnology Coordinated Infrastructure (NNCI) site at the University of Washington with partial support from the National Science Foundation via awards NNCI-1542101 and NNCI-2025489.

\section*{Author Contributions}

A.W.S., J.X., and M.C. contributed equally to this work. E.S. and A.M. conceived of the idea and provided funding. S.C. was involved in designing preliminary convolutional meta-optics and knowledge-distillation approaches. J.X. and E.S. developed the knowledge distillation approach. J.X. trained the neural network and analyzed the experiment data. A.W.S. developed the PSF-engineering approach and designed the meta-optics. M.C. fabricated the meta-optics and conducted the optical experiments. L.H. wrote the script to automate the experiment. J.F. advised on the fabrication and experiment. J.F., E.S., and A.M. provided project management and edited the manuscript.

\section*{Competing Interests}

A.M. is a co-founder of Tunoptix, which aims to commercialize meta-optics technology.

% \bibliographystyle{abbrvnat}

% \bibliography{Bibliography}% common bib file

\printbibliography[title={References}]  % Customize the title if necessary

\end{document}